\documentclass[10pt,twocolumn,letterpaper]{article}

\usepackage{cvpr}
\usepackage{times}
\usepackage{epsfig}
\usepackage{graphicx}
\usepackage{amsmath}
\usepackage{amssymb}
\usepackage{makecell}
\usepackage{multirow}
\usepackage{array}
\usepackage{color}
\usepackage{bm}
\usepackage{threeparttable}

\usepackage[pagebackref=true,breaklinks=true,letterpaper=true,colorlinks,bookmarks=false]{hyperref}

\newcommand{\urllink}{\fontsize{7.5pt}{\baselineskip}\selectfont}

\cvprfinalcopy 

\def\cvprPaperID{2493} 

\ifcvprfinal\fi
\begin{document}

\title{Relation-Shape Convolutional Neural Network for Point Cloud Analysis}

\author{Yongcheng Liu$^{\dag \hspace{0.5pt} \ddag}$ \quad\quad Bin Fan\thanks{Corresponding author: Bin Fan}\hspace{4pt}$^{\dag}$ \quad\quad Shiming Xiang$^{\dag \hspace{0.5pt} \ddag}$ \quad\quad Chunhong Pan$^{\dag}$\\
$^{\dag}$\hspace{1pt}National Laboratory of Pattern Recognition, Institute of Automation, Chinese Academy of Sciences\\
$^{\ddag}$\hspace{1pt}School of Artificial Intelligence, University of Chinese Academy of Sciences\\
{\tt\small Email:\;\{{yongcheng.liu,\;bfan,\;smxiang,\;chpan}\}@nlpr.ia.ac.cn}
}

\maketitle
\thispagestyle{empty}

\begin{abstract}
\label{abstract}
Point cloud analysis is very challenging, as the shape implied in irregular points is difficult to capture.
In this paper, we propose RS-CNN, namely, \textbf{R}elation-\textbf{S}hape \textbf{C}onvolutional \textbf{N}eural \textbf{N}etwork, which extends regular grid CNN to irregular configuration for point cloud analysis.
The key to RS-CNN is learning from relation, \textit{i.e.}, the geometric topology constraint among points.
Specifically, the convolutional weight for local point set is forced to learn a high-level relation expression from predefined geometric priors, between a sampled point from this point set and the others.
In this way, an inductive local representation with explicit reasoning about the spatial layout of points can be obtained, which leads to much shape awareness and robustness.
With this convolution as a basic operator, RS-CNN, a hierarchical architecture can be developed to achieve contextual shape-aware learning for point cloud analysis.
Extensive experiments on challenging benchmarks across three tasks verify RS-CNN achieves the state of the arts. 
\end{abstract}

\section{Introduction}
\label{sec:introduction}
\begin{figure}[t]
\centerline{\includegraphics[width=8.2cm]{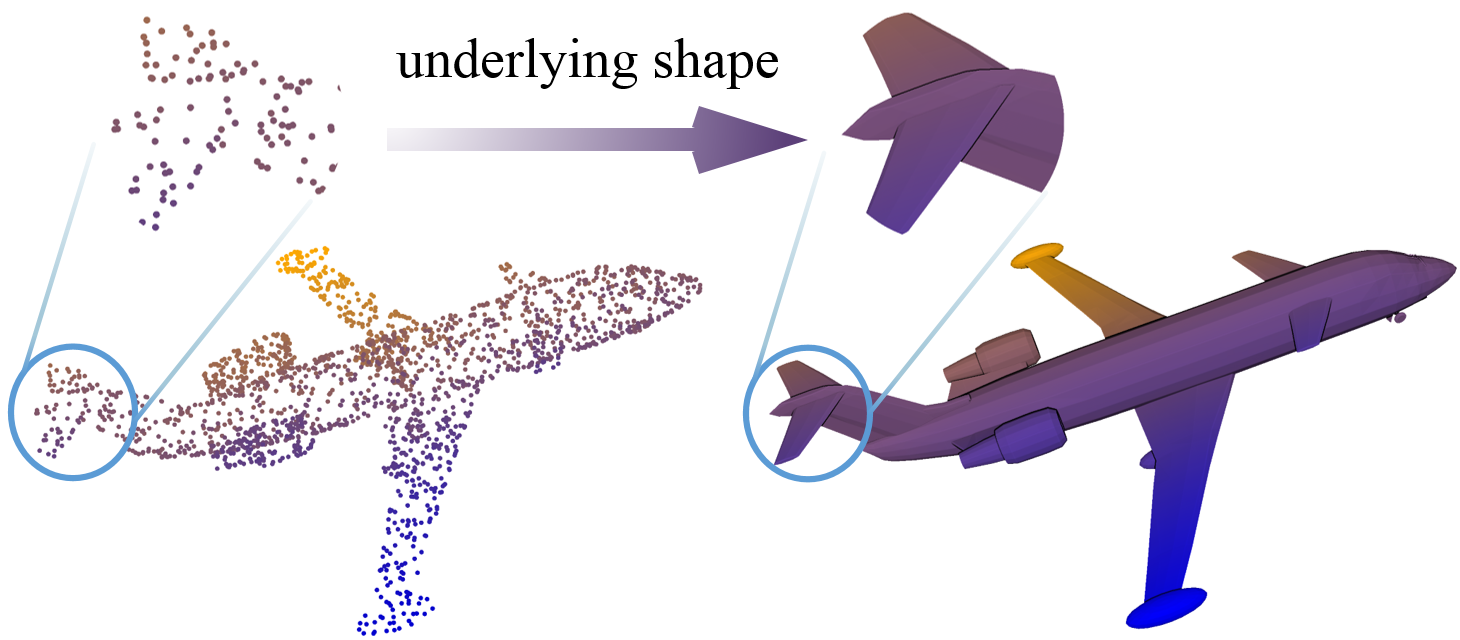}}
\caption{Left part: Point cloud. Right part: Underlying shape formed by this point cloud.}
\label{fig1:pc_vs_shape}
\end{figure}

Recently, the analysis of 3D point cloud has drawn a lot of attention, as it has many applications such as autonomous driving and robot manipulation.
However, this task is very challenging, since it is difficult to infer the underlying shape formed by these irregular points (see Fig.~\ref{fig1:pc_vs_shape} for detail).
%

For this issue, much effort is focused on replicating the remarkable success of convolutional neural network (CNN) on regular grid data (\textit{e.g.}, image) analysis~\cite{alexnet, VGG}, to irregular point cloud processing~\cite{c2_pointnet2,c23,c21,c5_generalcnn,c6_synccnn,c10_splatnet,c14_scn}.
Some works transform point cloud to regular voxels~\cite{modelnet40,vox2,c45} or multi-view images~\cite{multiview1,c37,multiview2} for easy application of classic grid CNN.
These transformations, however, usually lead to much loss of inherent geometric information in 3D point cloud, as well as high complexity.

To directly process point cloud, PointNet~\cite{c1_pointnet} independently learns on each point and gathers the final features for a global representation.
Though impressive, this design ignores local structures that have been proven to be important for abstracting high-level visual concepts in image CNN~\cite{visualize}.
To solve this problem, some works partition point cloud into several subsets by sampling~\cite{c2_pointnet2} or superpoint~\cite{c8_superpoint}.
Then a hierarchy is built to learn contextual representation from local to global.
Nevertheless, this extremely relies on effective inductive learning of local subsets, which is quite intractable to achieve.

Generally, there are mainly three challenges for learning from point set $P \subset \mathbb{R}^3$:
(1) $P$ is unordered, thus requiring the learned representation being permutation invariant;
(2) $P$ distributes in 3D geometric space, thus demanding the learned representation being robust to rigid transformation (\textit{e.g.}, rotation and translation);
(3) $P$ forms an underlying shape, therefore, the learned representation should be of discriminative shape awareness.
The issue (1) has been well resolved by symmetric function~\cite{c1_pointnet,c6_synccnn,c24}, while (2) and (3) still demand for a full exploration.
The goal of this work is to extend regular grid CNN to irregular configuration for handling these issues together.

To this end, we propose a relation-shape convolutional neural network (aliased as RS-CNN).
The key to RS-CNN is learning from relation, \textit{i.e.}, the geometric topology constraint among points, which in our view can encode meaningful shape information in 3D point cloud.

Specifically, each local convolutional neighborhood is constructed by taking a sampled point $x$ as the centroid and the surrounding points as its neighbors $\mathcal{N}(x)$.
Then, the convolutional weight is forced to learn a high-level relation expression from predefined geometric priors, \textit{i.e.}, intuitive low-level relation between $x$ and $\mathcal{N}(x)$.
%
%
By convoluting in this way, an inductive representation with explicit reasoning about the spatial layout of points can be obtained.
It discriminatively reflects the underlying shape that irregular points form thus is shape-aware.
Furthermore, it can benefit from geometric priors, including the invariance to points permutation and the robustness to rigid transformation (\textit{e.g.}, translation and rotation).
With this convolution as a basic operator, a hierarchical CNN-like architecture, \textit{i.e.}, RS-CNN, can be developed to achieve contextual shape-aware learning for point cloud analysis.

The key contributions are highlighted as follows:
\vspace{-3pt}
\begin{itemize}
\setlength{\itemsep}{0ex}
\item A novel learn-from-relation convolution operator called relation-shape convolution is proposed. It can explicitly encode geometric relation of points, thus resulting in much shape awareness and robustness;
\item A deep hierarchy equipped with the relation-shape convolution, \textit{i.e.}, RS-CNN, is proposed. It can extend regular grid CNN to irregular configuration for achieving contextual shape-aware learning of point cloud;
\item Extensive experiments on challenging benchmarks across three tasks, as well as thorough empirical and theoretical analysis, demonstrate RS-CNN achieves the state of the arts.
\end{itemize} 
\vspace{-11pt}

\section{Related Work}
\label{sec:related work}
\noindent \textbf{View-based and volumetric methods.}\,\,~View-based methods represent a 3D shape as a group of 2D views from different angles. Recently, many works~\cite{multiview1,c37,multiview2,multiview3,c49,c51} have been proposed to recognize these view images with deep neural networks. They often finetune a pre-trained image-based architecture for accurate recognition. However, 2D projections could cause loss of shape information due to self-occlusions, and it often demands a huge number of views for decent performance.

Volumetric methods convert the input 3D shape into a regular 3D grid, over which classic CNN can be employed~\cite{modelnet40,vox2,c45}. The main limitation is the quantization loss of the shape due to the low resolution enforced by 3D grid. Recent space partition methods like K-d trees~\cite{c26} or octrees~\cite{c29,vox3,c53} rescue some resolution issues but still rely on the subdivision of a bounding volume rather than a local geometric shape. In contrast to these methods, our work aims to process 3D point cloud directly.

\noindent \textbf{Deep learning on point cloud.}\,\,~PointNet~\cite{c1_pointnet} pioneers this route by independently learning on each point and gathering the final features with max pooling. Yet this design neglects local structures, which have been proven important for the success of CNN. To remedy this, PointNet++~\cite{c2_pointnet2} suggests a hierarchical application of PointNet to multiple subsets of point cloud. Local structure exploitation with PointNet is also investigated in~\cite{PCPNet, c9_kcnet}. In addition, Superpoint~\cite{c8_superpoint} is proposed to partition point cloud into geometric elements. Graph convolution network is applied on a local graph created by neighboring points~\cite{c14_scn,c19,c30}. However, these methods do not explicitly model the local spatial layout of points, thus acquiring less shape awareness. By contrast, our work captures the spatial layout of points by learning a high-level relation expression among points.

Some works map point cloud to a high-dimensional space to facilitate the application of classic CNN. SPLATNet~\cite{c10_splatnet} maps the input points onto a sparse lattice, then processing with bilateral convolution~\cite{bcl}. PCNN~\cite{c16_eocnn} extends the function over point cloud to a continuous volumetric function over ambient space. These methods could cause loss of geometric information, while our method directly operates on point cloud without introducing such loss.

Another key issue is the irregularity of points. Some works focus on analyzing symmetric functions that are equivariant to point sets learning~\cite{c1_pointnet,c6_synccnn,c24,c20}. Some other works~\cite{c1_pointnet,c27} develop alignment network for the robustness to rigid transformation in 3D space. However, the alignment learning is a suboptimal solution for this issue. Some traditional descriptors like Fast Point Feature Histograms can be invariant to translation and rotation, yet they are often less effective for high-level shape understanding. Our method that learns on geometric relation among points is naturally robust to rigid transformation, whilst being highly effective due to the powerfulness of deep nets.

\noindent \textbf{Relation learning.}\,\,~To learn a data-dependent weight from relation has been explored in the field of image and video analysis. Spatial transformer~\cite{stn} learns a transition matrix to align 2D images. Non-local network~\cite{non-local} learns long-term relation across video frames. Relation networks~\cite{relation_detection} learn position relation across objects. DFN~\cite{dfn_net} proposes general dynamic filters that inspire many subsequent works.

There are also some works focusing on the relation learning in 3D point cloud. DGCNN~\cite{c22} captures similar local shapes by learning point relation in a high-dimensional feature space, yet this relation could be unreliable in some cases. Wang \textit{et al.}~\cite{Param_conv} propose a parametric continuous convolution that is based on computable relation among points, but they do not explicitly learn from local to global like classic CNN. By contrast, our method learns a high-level relation expression from geometric priors in 3D space, and performs contextual local-to-global shape learning. 

\vspace{-3pt}

\section{Shape-Aware Representation Learning}
\label{sec:method}
\begin{figure*}[t]
\centerline{\includegraphics[width=17cm]{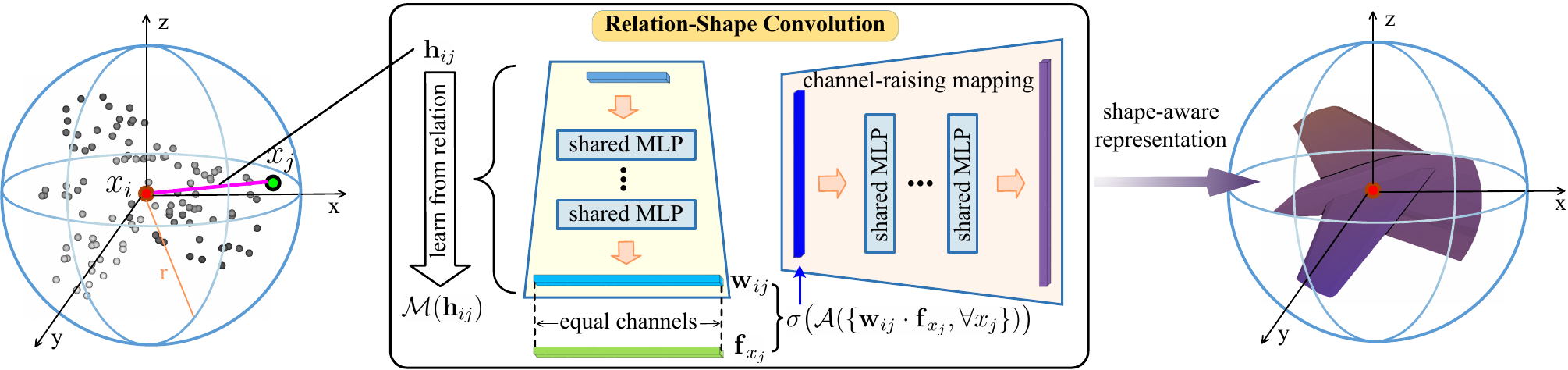}}
\caption{Overview of relation-shape convolution (RS-Conv). The key is to learn from relation. Specifically, the convolutional weight for $x_{j}$ is converted to ${\bm{\mathrm w}}_{ij}$, which learns a mapping $\mathcal{M}$ (Eq.~\eqref{Eq2:transform_relation}) on predefined geometric relation vector ${\bm{\mathrm h}}_{ij}$. In this way, the inductive convolutional representation $\sigma \big( \mathcal{A}(\{{\bm{\mathrm w}}_{ij} \cdot {\bm{\mathrm f}}_{x_j}, \hspace{0.1pt} \forall x_j\}) \big)$ (Eq.~\eqref{Eq3:graph_relation}) can expressively reason the spatial layout of points, resulting in discriminative shape awareness. As in image CNN~\cite{VGG}, further channel-raising mapping is conducted for a more powerful shape-aware representation.}
\label{fig2:RS-GC}
\end{figure*}

The core of point cloud analysis is to discriminatively represent the underlying shape with robustness. Here we learn contextual shape-aware representation for this goal, by extending regular grid CNN to irregular configuration with a novel relation-shape convolution (RS-Conv).

\subsection{Relation-Shape Convolution}
Local-to-global learning, which has gained remarkable success in image CNN~\cite{alexnet, VGG}, is a promising solution for contextual shape representation. However, it extremely relies on shape-aware inductive learning from irregular point subsets, which remains a quite intractable problem.

\vspace{6pt}
\noindent\textbf{Modeling.}\,\,~To overcome this issue, we model local point subset $P_{\text{sub}} \subset \mathbb{R}^3$ to be a spherical neighborhood, with a sampled point $x_{i}$ as the centroid and surrounding points as its neighbors $x_j \in \mathcal{N}(x_{i})$. The left-most part of Fig.~\ref{fig2:RS-GC} illustrates this modeling. Then, our goal is to learn an inductive representation ${\bm{\mathrm f}}_{P_{\text{sub}}}$ of this neighborhood, which should discriminatively encode the underlying shape information. To this end, we formulate a general convolutional operation as
\begin{equation}
\label{Eq1:graph_aggre}
{\bm{\mathrm f}}_{P_{\text{sub}}} = \sigma \big( \mathcal{A}(\{\mathcal{T}({\bm{\mathrm f}}_{x_j}), \ \forall x_j\}) \big)\footnote{In this paper, the bias term is omitted for clarity.}, \ d_{ij} < r \ \forall x_j \in \mathcal{N}(x_{i}),
\end{equation}
where $x$ is a 3D point and ${\bm{\mathrm f}}$ is a feature vector. $d_{ij}$ is the Euclidean distance between $x_{i}$ and $x_{j}$, and $r$ is the sphere radius. Here ${\bm{\mathrm f}}_{P_{\text{sub}}}$ is obtained by first transforming the features of all the points in $\mathcal{N}(x_{i})$ with function $\mathcal{T}$, and then aggregating them with function $\mathcal{A}$ followed by a nonlinear activator $\sigma$. In this formulation, the two functions $\mathcal{A}$ and $\mathcal{T}$ are the key to ${\bm{\mathrm f}}_{P_{\text{sub}}}$. That is, the permutation invariance of point set can be achieved only when $\mathcal{A}$ is symmetric (\textit{e.g.}, summation) and $\mathcal{T}$ is shared over each point in $\mathcal{N}(x_{i})$.

\vspace{6pt}
\noindent\textbf{Limitations of classic CNN.}\,\,~In classic CNN, $\mathcal{T}$ is implemented as $\mathcal{T}({\bm{\mathrm f}}_{x_j}) = {\bm{\mathrm w}}_j \cdot {\bm{\mathrm f}}_{x_j}$, where ${\bm{\mathrm w}}_j$ is learnable weight and ``$\cdot$'' denotes element-wise multiplication. There are mainly two limitations of this convolution when applied on point cloud: 1) ${\bm{\mathrm w}}_j$ is not shared over each point in $\mathcal{N}(x_{i})$, resulting in variance to point permutation and incapability to process irregular $P_{\text{sub}}$ (\textit{e.g.}, different number); 2) the gradient of ${\bm{\mathrm w}}_j$ in backpropagation is only relevant to the isolated point $x_{j}$, leading to an implicit learning strategy, which could not bring much shape awareness and robustness to ${\bm{\mathrm f}}_{P_{\text{sub}}}$. This issue can be partly alleviated by some techniques like performing various data augmentations or using lots of convolutional filters, yet they are suboptimal.

\vspace{6pt}
\noindent\textbf{Conversion: Learn from relation.}\,\,~We argue that the above limitations can be mitigated by learning from relation. In the neighborhood of 3D space, the geometric relation between $x_{i}$ and all its neighbors $\mathcal{N}(x_{i})$ is an explicit expression about the spatial layout of points, which further discriminatively reflects the underlying shape. To capture this relation, we replace ${\bm{\mathrm w}}_j$ in classical CNN with ${\bm{\mathrm w}}_{ij}$, which learns a mapping $\mathcal{M}$ of a relation vector ${\bm{\mathrm h}}_{ij}$, \textit{i.e.}, the predefined geometric priors between $x_{i}$ and $x_{j}$. We call ${\bm{\mathrm h}}_{ij}$ as low-level relation. This process can be described as
\begin{equation}
\label{Eq2:transform_relation}
\mathcal{T}({\bm{\mathrm f}}_{x_j}) = {\bm{\mathrm w}}_{ij} \cdot {\bm{\mathrm f}}_{x_j} = \mathcal{M}({\bm{\mathrm h}}_{ij}) \cdot {\bm{\mathrm f}}_{x_j}.
\end{equation}
The goal of mapping $\mathcal{M}$ is to abstract high-level relation expression between two points, which can encode their spatial layout. Here we implement $\mathcal{M}$ with a shared multi-layer perceptron (MLP) due to its powerful mapping ability. This process is illustrated in the middle part of Fig.~\ref{fig2:RS-GC}. In this way, ${\bm{\mathrm w}}_j$ is neatly converted to ${\bm{\mathrm w}}_{ij}$, whose gradient (determined by ${\bm{\mathrm h}_{ij}}$) is relevant to both $x_{i}$ and $x_{j}$. Meanwhile, $\mathcal{M}$ is exactly shared over all the points in $\mathcal{N}(x_{i})$, making it independent to the irregularity of points. It can also be robust to rigid transformation that will be clarified in Sec~\ref{subsec:3.2}.

As a consequence, ${\bm{\mathrm f}}_{P_{\text{sub}}}$ in Eq.~\eqref{Eq1:graph_aggre} becomes
\begin{equation}
\label{Eq3:graph_relation}
{\bm{\mathrm f}}_{P_{\text{sub}}} = \sigma \big( \mathcal{A}(\{\mathcal{M}({\bm{\mathrm h}}_{ij}) \cdot {\bm{\mathrm f}}_{x_j}, \ \forall x_j\}) \big).
\end{equation}
This convolutional representation, with all the relation between $x_{i}$ and $\mathcal{N}{(x_{i})}$ aggregated, can achieve explicit reasoning about the spatial layout of points, thus resulting in discriminative shape awareness. For geometric priors, one can use 3D Euclidean distance as an intuitive description of low-level relation ${\bm{\mathrm h}}_{ij}$. Moreover, ${\bm{\mathrm h}}_{ij}$ can also be defined flexibly since $\mathcal{M}$ can map it to a high-dimensional relation vector for channel alignment with ${\bm{\mathrm f}}_{x_j}$ for easy multiplication. We will discuss ${\bm{\mathrm h}}_{ij}$ in detail in the experiment section.

\vspace{6pt}
\noindent\textbf{Channel-raising mapping.}\,\,~In Eq.~\eqref{Eq3:graph_relation}, the channel number of ${\bm{\mathrm f}}_{P_{\text{sub}}}$ is the same as the input feature ${\bm{\mathrm f}}_{x_j}$. This is inconsistent with classic image CNN that increases channel number while decreasing image resolution for a more abstract representation. For example, the channel number of 64-128-256-512 is set in VGG network~\cite{VGG}. Accordingly, we add a shared MLP on ${\bm{\mathrm f}}_{P_{\text{sub}}}$ for further channel-raising mapping. It is illustrated in the middle part of Fig.~\ref{fig2:RS-GC}.

\subsection{Properties}
\label{subsec:3.2}
RS-Conv in Eq.~\eqref{Eq3:graph_relation} can maintain four decent properties:

\vspace{6pt}
\noindent\textbf{Permutation invariance.}\,\,~In the inner mapping function $\mathcal{M}({\bm{\mathrm h}})$, both the low-level relation ${\bm{\mathrm h}}$ and the shared MLP $\mathcal{M}$ are invariant to the input order of points. Therefore, with the outer aggregation function $\mathcal{A}$ being symmetric, the permutation invariance can be satisfied.

\vspace{6pt}
\noindent \textbf{Robustness to rigid transformation.}\,\,~This property is well held in the high-level relation encoding $\mathcal{M}({\bm{\mathrm h}})$. It can be robust to rigid transformation, \textit{e.g.}, translation and rotation, when a suitable ${\bm{\mathrm h}}$ (\textit{e.g.}, 3D Euclidean distance) is defined.

\vspace{6pt}
\noindent \textbf{Points interaction.}\,\,~Points are not isolated and nearby points form a meaningful shape in geometric space. Thus their inherent interaction is critical for discriminative shape awareness. Our solution of relation learning explicitly encode the geometric relation among points, naturally capturing the interaction of points.

\vspace{6pt}
\noindent \textbf{Weight sharing.}\,\,~This is the key property that allows applying the same learning function over different irregular point subsets for robustness, as well as low complexity. In Eq.~\eqref{Eq3:graph_relation}, the symmetric $\mathcal{A}$, the shared MLP $\mathcal{M}$ and the predefined geometric priors ${\bm{\mathrm h}}$ are all independent to the irregularity of points. Hence, this property is also satisfied.

\subsection{Revisiting 2D Grid Convolution}
\begin{figure}[t]
\centerline{\includegraphics[width=8.2cm]{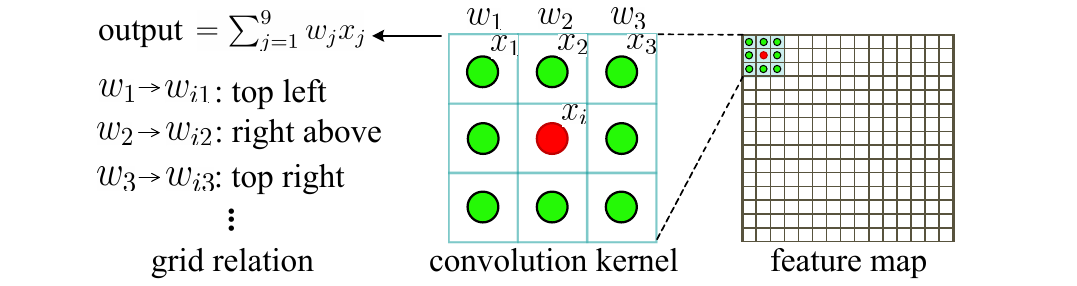}}
\caption{Illustration of 2D grid convolution with a kernel of $3 \times 3$.}
\label{fig3:2dconv_cropped}
\end{figure}

The proposed RS-Conv is a generic formulation of 2D grid convolution for relation reasoning. We clarify this with a neighborhood (convolution kernel) of $3 \times 3$ on a 2D-grid feature map, as illustrated in Fig.~\ref{fig3:2dconv_cropped}. Specifically, the summation function $\sum$ is a specific instance of the aggregation function $\mathcal{A}$. Moreover, note that $w_{j}$ always implies a fixed positional relation between $x_{i}$ and its neighbor $x_{j}$ in the regular grid. For example, $w_{1}$ always implies the top-left relation with $x_{i}$, and $w_{2}$ implies the right-above relation with $x_{i}$. In other words, $w_{j}$ is actually constrained to encode one kind of regular grid relation in the learning process. Therefore, our RS-Conv with relation learning is more general and can be applied to model 2D grid spatial relationship.

\subsection{RS-CNN for Point Cloud Analysis}
Using RS-Conv (Fig.~\ref{fig2:RS-GC}) as a basic operator and adopting a uniform sampling strategy, a hierarchical shape-aware learning architecture like classic CNN, namely, RS-CNN, can be developed for point cloud analysis as
\begin{equation}
\label{Eq4:single_ShapeGraph}
\mathbf{F}^{\ell}_{P_{N_{\ell}}} = \textsc{RS-Conv}(\mathbf{F}^{\ell-1}_{P_{N_{\ell-1}}}),
\end{equation}
where $\mathbf{F}^{\ell}_{P_{N_{\ell}}}$, features in layer $\ell$ of the sampled point set $P_{N_{\ell}}$ with number $N_{\ell}$, are obtained by applying RS-Conv on the features in the previous layer $\ell-1$.

\begin{figure}[t]
\centerline{\includegraphics[width=8.2cm]{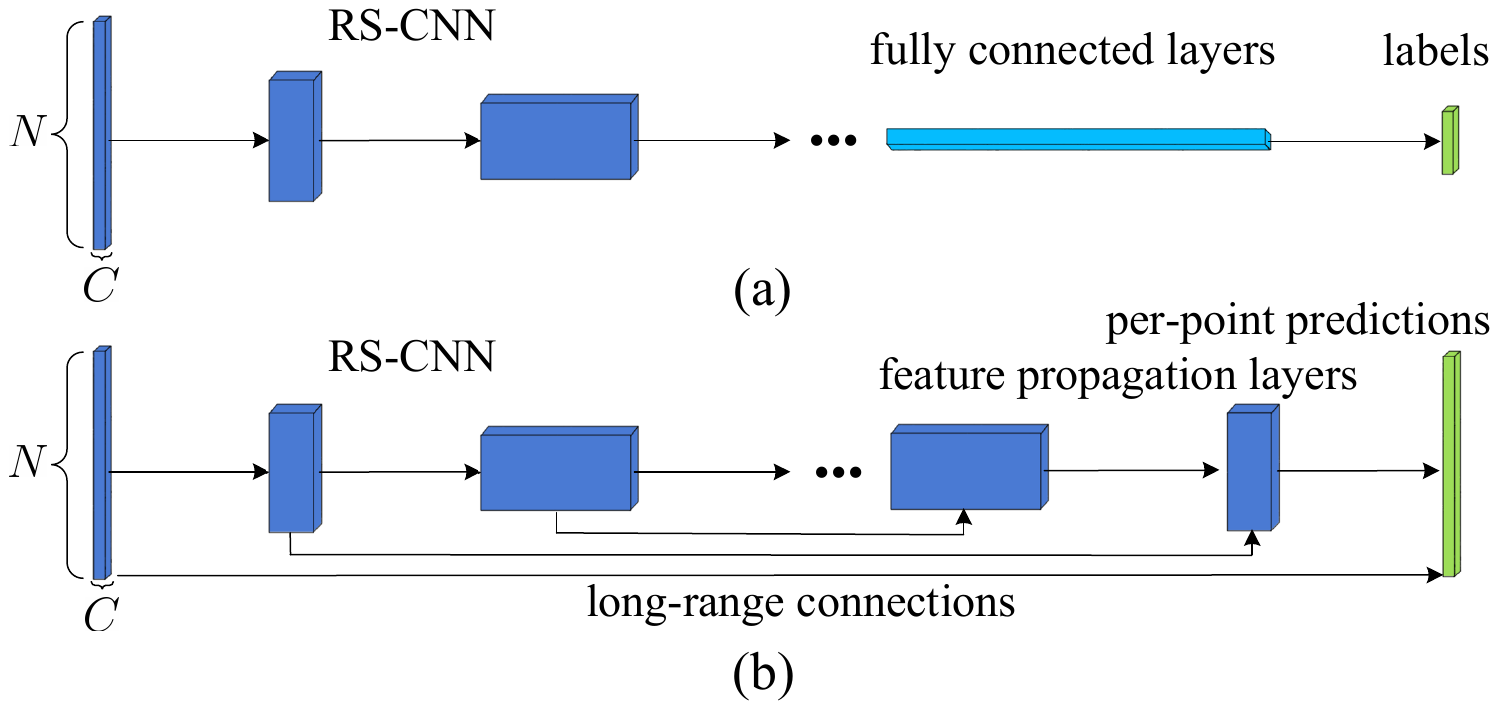}}
\caption{The architectures of RS-CNN applied in the classification (a) and segmentation (b) of point cloud. $N$ is the number of points and $C$ is the channel number.}
\label{fig4:cls_seg_net}
\end{figure}

Our RS-CNN applied in the classification and segmentation of point cloud is illustrated in Fig.~\ref{fig4:cls_seg_net}. In both tasks, RS-CNN is used for learning a group of hierarchical shape-aware representation. The final global representation followed by three fully connected (FC) layers is configured for classification. For segmentation, the learned multi-level representation is successively upsampled by feature propagation~\cite{c2_pointnet2} to generate per-point predictions. Both of them can be trained in an end-to-end manner.

\subsection{Implementation Details}
\label{subsec:3.5}
\noindent\textbf{RS-Conv in Eq.~\eqref{Eq3:graph_relation}.}\,\,~Symmetric function max pooling is applied as aggregation function $\mathcal{A}$. ReLU~\cite{relu} is used as nonlinear activator $\sigma$. For mapping function $\mathcal{M}$, a three-layer shared MLP is deployed since theoretically it can fit arbitrary continuous mappings~\cite{c50}. Low-level relation ${\bm{\mathrm h}}_{ij}$ is defined as a compact vector with 10 channels, \textit{i.e.}, (3D Euclidean distance, $x_{i}-x_{j}$, $x_{i}$, $x_{j}$). The channel-raising mapping is achieved by a single-layer shared MLP. Batch normalization~\cite{BN} is applied in each MLP.

\vspace{6pt}
\noindent\textbf{RS-CNN for points analysis.}\,\,~The farthest points are picked from point cloud for sampling local subsets to perform RS-Conv. In each neighborhood, a fixed number of neighbors are randomly sampled for batch processing, and they are normalized to take the centroid as the origin. To capture more sufficient geometric relation, we force RS-CNN to learn over three-scale neighborhoods centered on a sampled point with a shared weight. This is different from multi-scale grouping (MSG)~\cite{c2_pointnet2} that learns multi-scale features using multiple groups of weight. RS-CNN with 3 layers and 4 layers is deployed for classification and segmentation, respectively. Note that only 3D coordinates $\mathrm{xyz}$ are used as the input features to RS-CNN.

Our RS-CNN is implemented using Pytorch\footnote{\urllink \texttt{https://github.com/Yochengliu/Relation-Shape-CNN}}. The Adam optimization algorithm is employed for training, with a mini-batch size of $32$. The momentum for BN starts with $0.9$ and decays with a rate of $0.5$ every $20$ epochs. The learning rate begins with $0.001$ and decays with a rate of $0.7$ every $20$ epochs. The weight of RS-CNN is initialized using the techniques introduced by He \textit{et al}.~\cite{conf_iccv_HeZRS15}. 

\section{Experiment}
\label{sec:experiment}
In this section, we arrange comprehensive experiments to validate the proposed RS-CNN. First, we evaluate RS-CNN for point cloud analysis on three tasks (Sec~\ref{subsec4.1}). We then provide detailed experiments to carefully study RS-CNN (Sec~\ref{subsec4.2}). Finally, we visualize the shape features that RS-CNN captures and analyze the complexity (Sec~\ref{subsec4.3}).

\subsection{Point Cloud Analysis}
\label{subsec4.1}
\noindent \textbf{Shape classification.}\,\,~We evaluate RS-CNN on ModelNet40 classification benchmark~\cite{modelnet40}. It is composed of 9843 train models and 2468 test models in 40 classes. The point cloud data is sampled from these models by~\cite{c1_pointnet}. We uniformly sample 1024 points and normalize them to a unit sphere. During training, we augment the input data with random anisotropic scaling in the range [-0.66, 1.5] and translation in the range [-0.2, 0.2], as in~\cite{c26}. Meanwhile, dropout technique~\cite{dropout} with 50\% ratio is applied in FC layers. During testing, similar to~\cite{c1_pointnet,c2_pointnet2}, we perform ten voting tests with random scaling and average the predictions.

The quantitative comparisons with the state-of-the-art point-based methods are summarized in Table~\ref{Tab1:cls}, where RS-CNN outperforms all the $\mathrm{xyz}$-input methods. Specifically, RS-CNN reduces the error rate of PointNet++~\cite{c2_pointnet2} by 31.2\%, and surpasses its advanced version that uses additional normal data as well as very dense points (5k). Moreover, even using only $\mathrm{xyz}$ as the input, RS-CNN can also achieve a superior result (93.6\%) compared with the best additional-input method SO-Net~\cite{c20} (93.4\%). This convincingly verifies the effectiveness of our RS-CNN.

\begin{figure}[t]
\centerline{\includegraphics[width=8.2cm]{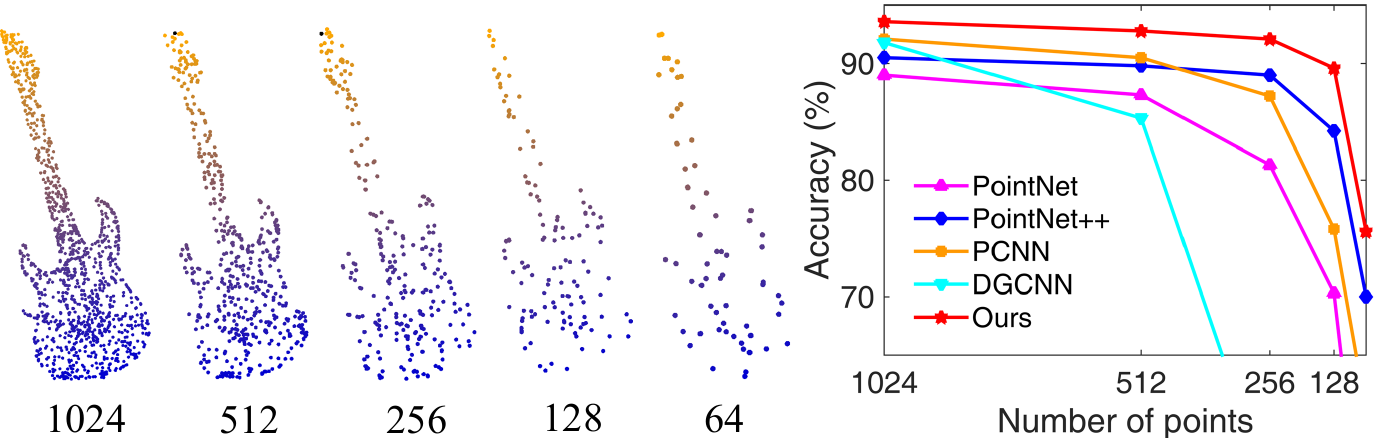}}
\caption{Left part: Point cloud with random point dropout. Right part: Test results of using sparser points as the input to a model trained with 1024 points.}
\label{fig5:sample_density}
\vspace{5pt}
\end{figure}

\begin{table}[t]
  \centering
  \small
  \caption{Shape classification results (\%) on ModelNet40 benchmark (nor: normal, ``-'': unknown).}
  \begin{tabular}{l|ccc}
  \Xhline{0.8pt}
  method & input & \#points & acc. \\
  \Xhline{0.5pt}
  Pointwise-CNN~\cite{c17_pointwise} & $\mathrm{xyz}$ & 1k & 86.1 \\
  Deep Sets~\cite{c24} & $\mathrm{xyz}$ & 1k & 87.1 \\
  ECC~\cite{c32} & $\mathrm{xyz}$ & 1k & 87.4 \\
  PointNet~\cite{c1_pointnet} & $\mathrm{xyz}$ & 1k & 89.2 \\
  SCN~\cite{c7_attsp} & $\mathrm{xyz}$ & 1k & 90.0 \\
  Flex-Conv~\cite{flex_conv} & $\mathrm{xyz}$ & 1k & 90.2 \\
  Kd-Net(depth=10)~\cite{c26} & $\mathrm{xyz}$ & 1k & 90.6 \\
  PointNet++~\cite{c2_pointnet2} & $\mathrm{xyz}$ & 1k & 90.7 \\
  KCNet~\cite{c9_kcnet} & $\mathrm{xyz}$ & 1k & 91.0 \\
  MRTNet~\cite{c45} & $\mathrm{xyz}$ & 1k & 91.2 \\
  Spec-GCN~\cite{c14_scn} & $\mathrm{xyz}$ & 1k & 91.5 \\
  PointCNN~\cite{c27} & $\mathrm{xyz}$ & 1k & 91.7 \\
  DGCNN~\cite{c22} & $\mathrm{xyz}$ & 1k & 92.2 \\
  PCNN~\cite{c16_eocnn} & $\mathrm{xyz}$ & 1k & 92.3 \\
  \textbf{Ours} & $\bm {\mathrm{xyz}}$ & \textbf{1k} & \textbf{93.6} \\
  SO-Net~\cite{c20} & $\mathrm{xyz}$ & 2k & 90.9 \\
  Kd-Net(depth=15)~\cite{c26} & $\mathrm{xyz}$ & 32k & 91.8 \\
  \Xhline{0.5pt}
  O-CNN~\cite{c29} & $\mathrm{xyz}$, nor & - & 90.6 \\
  Spec-GCN~\cite{c14_scn} & $\mathrm{xyz}$, nor & 1k & 91.8 \\
  PointNet++~\cite{c2_pointnet2} & $\mathrm{xyz}$, nor & 5k & 91.9 \\
  SpiderCNN~\cite{c21} & $\mathrm{xyz}$, nor & 5k & 92.4 \\
  SO-Net~\cite{c20} & $\mathrm{xyz}$, nor & 5k & 93.4 \\
  \Xhline{0.8pt}
  \end{tabular}
  \label{Tab1:cls}
\end{table}

We test the robustness of RS-CNN on sampling density, by using sparser points of number 1024, 512, 256, 128 and 64 as the input to a model trained with 1024 points. As in~\cite{c2_pointnet2}, random input dropout technique is applied for a fair comparison. Fig.~\ref{fig5:sample_density} shows the test results, where the compared methods are PointNet~\cite{c1_pointnet}, PointNet++~\cite{c2_pointnet2}, PCNN~\cite{c16_eocnn} and DGCNN~\cite{c22}. As can be seen, it is more difficult for shape recognition when points get sparser. Even so, RS-CNN is still considerably robust. It achieves nearly consistent robustness as PointNet++, whilst showing superior performance on each density.

\begin{table*}[t]
  \centering
  \caption{Shape part segmentation results (\%) on ShapeNet part benchmark (nor: normal, ``-'': unknown).}
  \footnotesize
  \begin{tabular}{p{1.8cm}|p{0.7cm}|p{0.5cm}|p{0.75cm}|p{0.32cm}p{0.32cm}p{0.32cm}p{0.32cm}p{0.32cm}p{0.32cm}p{0.32cm}p{0.32cm}p{0.32cm}p{0.32cm}p{0.32cm}p{0.32cm}p{0.32cm}p{0.32cm}p{0.32cm}p{0.32cm}}
  \Xhline{0.8pt}
  method & input & class mIoU & instance mIoU & air plane & bag & cap & car & chair & ear phone & guitar & knife & lamp & laptop & motor bike & mug & pistol & rocket & skate board & table \\
  \Xhline{0.5pt}
  Kd-Net~\cite{c26} & 4k & 77.4 & 82.3 & 80.1 & 74.6 & 74.3 & 70.3 & 88.6 & 73.5 & 90.2 & 87.2 & 81.0 & 94.9 & 57.4 & 86.7 & 78.1 & 51.8 & 69.9 & 80.3 \\
  PointNet~\cite{c1_pointnet} & 2k & 80.4 & 83.7 & 83.4 & 78.7 & 82.5 & 74.9 & 89.6 & 73.0 & 91.5 & 85.9 & 80.8 & 95.3 & 65.2 & 93.0 & 81.2 & 57.9 & 72.8 & 80.6 \\
  RS-Net~\cite{C28} & - & 81.4 & 84.9 & 82.7 & \textbf{86.4} & 84.1 & 78.2 & 90.4 & 69.3 & 91.4 & 87.0 & 83.5 & 95.4 & 66.0 & 92.6 & 81.8 & 56.1 & 75.8 & 82.2 \\
  SCN~\cite{c7_attsp} & 1k & 81.8 & 84.6 & 83.8 & 80.8 & 83.5 & 79.3 & 90.5 & 69.8 & \textbf{91.7} & 86.5 & 82.9 & 96.0 & 69.2 & 93.8 & 82.5 & \textbf{62.9} & 74.4 & 80.8 \\
  PCNN~\cite{c16_eocnn} & 2k & 81.8 & 85.1 & 82.4 & 80.1 & 85.5 & 79.5 & 90.8 & 73.2 & 91.3 & 86.0 & 85.0 & 95.7 & 73.2 & 94.8 & 83.3 & 51.0 & 75.0 & 81.8 \\
  SPLATNet~\cite{c10_splatnet} & - & 82.0 & 84.6 & 81.9 & 83.9 & 88.6 & 79.5 & 90.1 & 73.5 & 91.3 & 84.7 & 84.5 & \textbf{96.3} & 69.7 & \textbf{95.0} & 81.7 & 59.2 & 70.4 & 81.3 \\
  KCNet~\cite{c9_kcnet} & 2k & 82.2 & 84.7 & 82.8 & 81.5 & 86.4 & 77.6 & 90.3 & 76.8 & 91.0 & 87.2 & 84.5 & 95.5 & 69.2 & 94.4 & 81.6 & 60.1 & 75.2 & 81.3 \\
  DGCNN~\cite{c22} & 2k & 82.3 & 85.1 & \textbf{84.2} & 83.7 & 84.4 & 77.1 & 90.9 & 78.5 & 91.5 & 87.3 & 82.9 & 96.0 & 67.8 & 93.3 & 82.6 & 59.7 & 75.5 & 82.0 \\
  \textbf{Ours} & \textbf{2k} & \textbf{84.0} & \textbf{86.2} & 83.5 & 84.8 & \textbf{88.8} & \textbf{79.6} & \textbf{91.2} & \textbf{81.1} & 91.6 & \textbf{88.4} & \textbf{86.0} & 96.0 & \textbf{73.7} & 94.1 & \textbf{83.4} & 60.5 & \textbf{77.7} & \textbf{83.6} \\
  \Xhline{0.5pt}
  \scriptsize{PointNet++}~\cite{c2_pointnet2} & 2k,nor & 81.9 & 85.1 & 82.4 & 79.0 & 87.7 & 77.3 & 90.8 & 71.8 & 91.0 & 85.9 & 83.7 & 95.3 & 71.6 & 94.1 & 81.3 & 58.7 & 76.4 & 82.6 \\
  SyncCNN~\cite{c3_synccnn} & mesh & 82.0 & 84.7 & 81.6 & 81.7 & 81.9 & 75.2 & 90.2 & 74.9 & 93.0 & 86.1 & 84.7 & 95.6 & 66.7 & 92.7 & 81.6 & 60.6 & 82.9 & 82.1 \\
  SO-Net~\cite{c20} & 1k,nor & 80.8 & 84.6 & 81.9 & 83.5 & 84.8 & 78.1 & 90.8 & 72.2 & 90.1 & 83.6 & 82.3 & 95.2 & 69.3 & 94.2 & 80.0 & 51.6 & 72.1 & 82.6 \\
  \scriptsize{SpiderCNN}~\cite{c21} & 2k,nor & 82.4 & 85.3 & 83.5 & 81.0 & 87.2 & 77.5 & 90.7 & 76.8 & 91.1 & 87.3 & 83.3 & 95.8 & 70.2 & 93.5 & 82.7 & 59.7 & 75.8 & 82.8 \\
  \Xhline{0.8pt}
  \end{tabular}
  \label{Tab2:seg}
\end{table*}

\vspace{10pt}
\noindent \textbf{Shape part segmentation.}\,\,~Part segmentation is a challenging task for fine-grained shape analysis. We evaluate RS-CNN for this task on ShapeNet part benchmark~\cite{c54} and follow the data split in~\cite{c1_pointnet}. This dataset contains 16881 shapes with 16 categories, and is labeled in 50 parts in total. As in~\cite{c1_pointnet}, we randomly pick 2048 points as the input and concatenate the one-hot encoding of the object label to the last feature layer. During testing, we also apply ten voting tests using random scaling. Except for standard IoU (\textit{Inter-over-Union}) on each category, we also report two types of mean IoU (mIoU) that are averaged across all classes and all instances, respectively.

\begin{figure}[t]
\centerline{\includegraphics[width=8cm]{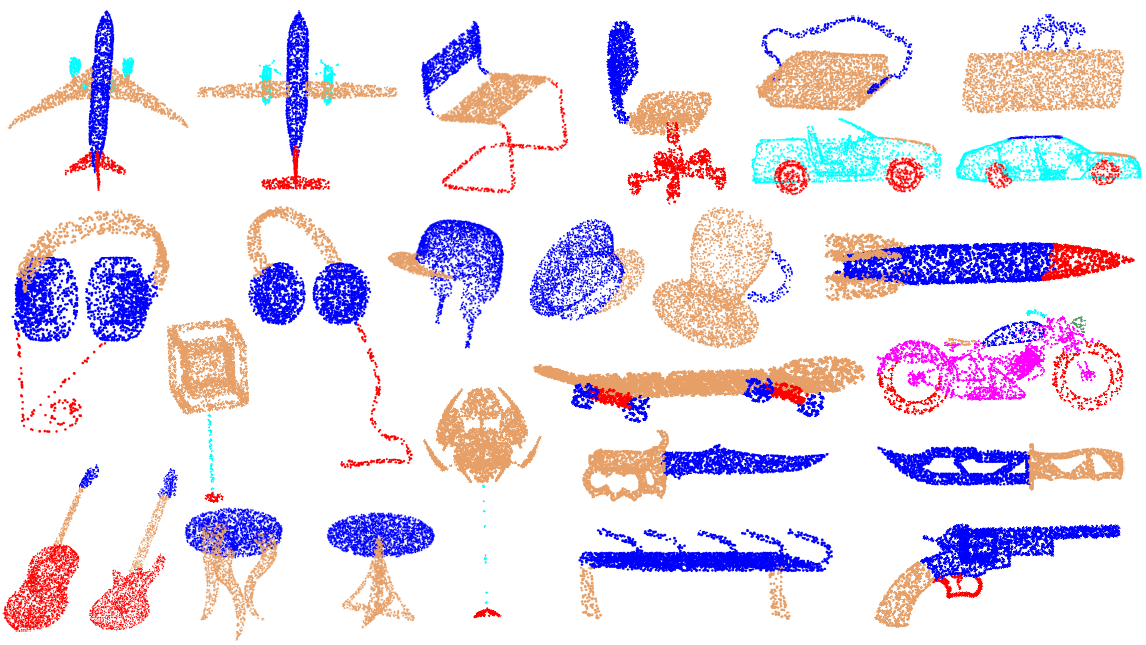}}
\caption{Segmentation examples on ShapeNet part benchmark.}
\label{fig6:part_show}
\vspace{-10pt}
\end{figure}

Table~\ref{Tab2:seg} summarizes the quantitative comparisons with the state-of-the-art methods, where RS-CNN achieves the best performance with class mIoU of 84.0\% and instance mIoU of 86.2\%. This considerably surpasses the second best $\mathrm{xyz}$-based methods, \textit{i.e.}, DGCNN~\cite{c22} with 82.3\% (1.7$\uparrow$) in class mIoU and PCNN~\cite{c16_eocnn} with 85.1\% (1.1$\uparrow$) in instance mIoU, respectively. Noticeably, RS-CNN sets new state of the arts in the $\mathrm{xyz}$-based methods over ten categories. These improvements demonstrate the robustness of RS-CNN to diverse shape structures. Fig.~\ref{fig6:part_show} shows some segmentation examples. One can see that although the part shapes implied in irregular points are varied and they may be very confusing to recognize, RS-CNN can also segment them out with decent accuracy.

\vspace{6pt}
\noindent \textbf{Normal estimation.}\,\,~Normal estimation in point cloud is a crucial step for numerous applications, such as surface reconstruction and rendering. This task is very challenging since it requires a higher level of reasoning, which goes beyond the underlying shape recognition. We take normal estimation as a supervised regression task, and achieve it using the segmentation network. The cosine-loss between the normalized output and ground truth normal is applied for regression training. ModelNet40 dataset is used for evaluation, with uniformly sampled 1024 points as the input.

\begin{table}[t]
  \centering
  \caption{Normal estimation error on ModelNet40 dataset.}
  \begin{tabular}{l|ccc}
  \Xhline{0.8pt}
  dataset & method & \#points & error \\
  \Xhline{0.5pt}
  ModelNet40 & PointNet~\cite{c16_eocnn} & 1k & 0.47 \\
                     & PointNet++~\cite{c16_eocnn} & 1k & 0.29 \\
                     & PCNN~\cite{c16_eocnn} & 1k & 0.19 \\
                     & \textbf{Ours} & \textbf{1k} & \textbf{0.15} \\
  \Xhline{0.8pt}
  \end{tabular}
  \label{Tab3:normal}
  \vspace{-5pt}
\end{table}

\begin{figure}[t]
\centerline{\includegraphics[width=8.2cm]{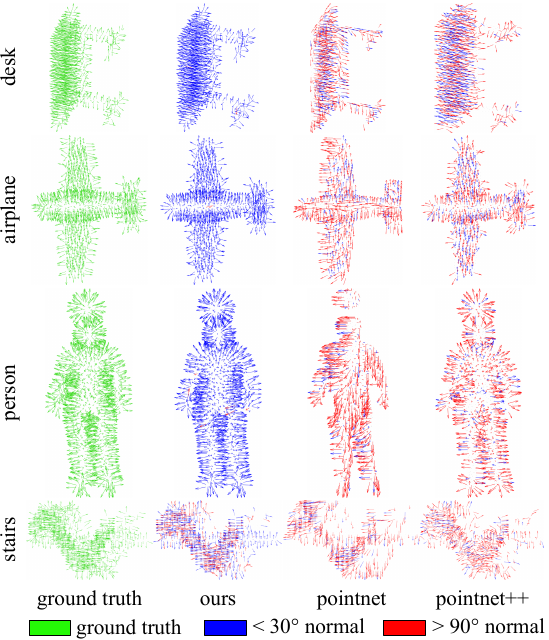}}
\caption{Normal estimation on ModelNet40 dataset. For clearness, we only show predictions with angle less than 30$^{\circ}$ in \textcolor[rgb]{0.00,0.00,1.00}{blue}, and angle greater than 90$^{\circ}$ in \textcolor[rgb]{1.00,0.00,0.00}{red} between ground truth normals.}
\label{fig7:normal_show}
\vspace{-5pt}
\end{figure}

The quantitative results are summarized in Table~\ref{Tab3:normal}. RS-CNN outperforms other advanced methods on this task with a lower error of 0.15. This significantly reduces the error of PointNet++ (0.29) by 48.3\%. Fig.~\ref{fig7:normal_show} shows some normal estimation examples, where our RS-CNN with geometric relation learning can obtain more decent predictions. However, RS-CNN could also be less effective for some intractable shapes, such as spiral stairs and intricate plants.

\subsection{RS-CNN Design Analysis}
\label{subsec4.2}
In this section, we first perform a detailed ablation study on RS-CNN. Then, we discuss the choices of aggregation function $\mathcal{A}$, mapping function $\mathcal{M}$ and low-level relation ${\bm{\mathrm h}}$ in Eq.~\eqref{Eq3:graph_relation}. Finally, we validate the robustness of RS-CNN on point permutation and rigid transformation. All experiments are conducted on ModelNet40 classification dataset.

\vspace{12pt}
\noindent \textbf{Ablation study.}\,\,~The results are summarized in Table~\ref{Tab4:ablation}. The baseline (model A) is set to learn without geometric relation encoding, but with a shared three-layer MLP as feature transformation function $\mathcal{T}$ in Eq.~\eqref{Eq1:graph_aggre}.

The baseline only gets an accuracy of 87.2\%. Yet with geometric relation learning, it is significantly improved to 89.9\% (model B). This convincingly verifies the effectiveness of our RS-CNN. Then, a great improvement of 2\% is gained after using BN (model C), maybe because it can greatly ease the network training. Moreover, dropout technique improves the result by 0.3\% (model D). As mentioned in Sec~\ref{subsec:3.5}, RS-CNN should be able to benefit from sufficient geometric relation. This is verified by model E (92.5\%) and model F (92.9\%) that perform two-scale and three-scale relation learning, respectively. Eventually, with ten voting tests, an impressive accuracy of 93.6\% (model G) can be obtained with only $\mathrm{xyz}$ features.

\begin{table}[t]
  \centering
  \footnotesize
  \caption{Ablation study of RS-CNN (\%). ``DP'' indicates the dropout technique in FC layers of the classification network.}
  \begin{tabular}{c|cccccc|c}
  \Xhline{0.8pt}
  model & \#points & relation & BN & DP & scale & voting & acc. \\
  \Xhline{0.5pt}
  A & 1k &  &  &  & 1 &  & 87.2 \\
  B & 1k & $\checkmark$ &  &  & 1 &  & 89.9 \\
  C & 1k & $\checkmark$ & $\checkmark$ &  & 1 &  & 91.9 \\
  D & 1k & $\checkmark$ & $\checkmark$ & $\checkmark$ & 1 &  & 92.2 \\
  E & 1k & $\checkmark$ & $\checkmark$ & $\checkmark$ & 2 &  & 92.5  \\
  F & 1k & $\checkmark$ & $\checkmark$ & $\checkmark$ & 3 &  & 92.9 \\
  G & 1k & $\checkmark$ & $\checkmark$ & $\checkmark$ & 3 & $\checkmark$ & \textbf{93.6} \\
  H & 2k & $\checkmark$ & $\checkmark$ & $\checkmark$ & 3 & $\checkmark$ & \textbf{93.6} \\
  I & 1k &  & $\checkmark$ & $\checkmark$ & 3 & $\checkmark$ & 90.1 \\
  \Xhline{0.8pt}
  \end{tabular}
  \label{Tab4:ablation}
\end{table}

To investigate the impact of the number of input points on RS-CNN, we also train the network with 2048 points but find no improvement (model H). In addition, to compare with the baseline (model A) more fairly, we set a new baseline (model I) that works with all the techniques but relation learning. It gets an accuracy of 90.1\%, which RS-CNN can also surpass by 3.5\%. We speculate that RS-CNN with geometric relation reasoning can acquire more discriminative shape awareness, and this awareness can be greatly enhanced by multi-scale relation learning.

\vspace{12pt}
\noindent \textbf{Aggregation function $\mathcal{A}$.}\,\,~Three symmetric functions: max pooling (max), average pooling (avg.) and summation (sum), are employed to study the effect of $\mathcal{A}$ on RS-CNN. Table~\ref{Tab5:inner_out_func} summarizes the results. As can be seen, with $\mathcal{M}$ using three layers, max pooling achieves the best performance while average pooling and summation get the same accuracy. The reason may be that max pooling can select the biggest feature response, thus keeping the most expressive representation and removing redundant information.

\vspace{12pt}
\noindent \textbf{Mapping function $\mathcal{M}$.}\,\,~The results of $\mathcal{M}$ deployed with different layers are summarized in the first three rows of Table~\ref{Tab5:inner_out_func}. One can see that the best accuracy of 93.6\% is obtained by a shared three-layer MLP, and it decreases by 0.9\% when increasing the number of layers. The reason might be that $\mathcal{M}$ with four layers brings some difficulty for network training. Noticeably, RS-CNN can also get a decent accuracy of 92.4\% with $\mathcal{M}$ using only two layers. This verifies the powerfulness of relation learning for underlying shape capturing from point cloud.

\begin{table}[t]
  \centering
  \caption{The results (\%) of different designs on aggregation function $\mathcal{A}$ and mapping function $\mathcal{M}$ (Eq.~\eqref{Eq3:graph_relation}) ($\mathcal{M}_{(k)}$: $k$-layer MLP).}
  \begin{tabular}{c|ccc|c}
  \Xhline{0.8pt}
  $\mathcal{A}$ & $\mathcal{M}_{(2)}$ & $\mathcal{M}_{(3)}$ & $\mathcal{M}_{(4)}$ & acc. \\
  \Xhline{0.5pt}
  max & $\checkmark$ &  &  & 92.4 \\
  max & & $\checkmark$ &  & \textbf{93.6} \\
  max &  &  & $\checkmark$ & 92.7 \\
  avg. &  & $\checkmark$ &  & 91.6 \\
  sum &  & $\checkmark$ &  & 91.6 \\
  \Xhline{0.8pt}
  \end{tabular}
  \label{Tab5:inner_out_func}
  \vspace{10pt}
\end{table}

\begin{table}[t]
  \centering
  \small
  \caption{The results (\%) of five intuitive low-level relations ${\bm{\mathrm h}}$ (Ed: Euclidean distance, cosd: cosine distance, $x^{\text{nor}}$: normal of $x$, $x'$: 2D projection of $x$). Model A applies only 3D Euclidean distance as ${\bm{\mathrm h}}$; Model B adds the coordinates difference to model A; Model C adds the coordinates of two points to model B; Model D utilizes the normals of two points and their cosine distance as ${\bm{\mathrm h}}$; Model E projects 3D points onto a 2D plane of $\mathrm{XY}$, $\mathrm{XZ}$ and $\mathrm{YZ}$.}
  \begin{tabular}{c|lc|c}
  \Xhline{0.8pt}
  model & low-level relation ${\bm{\mathrm h}}$ & channels & acc. \\
  \Xhline{0.5pt}
  A & (3D-Ed) & 1 & 92.5 \\
  B & (3D-Ed, $x_{i} - x_{j}$) & 4 & 93.0  \\
  C & (3D-Ed, $x_{i} - x_{j}$, $x_{i}$, $x_{j}$) & 10 & \textbf{93.6} \\
  D & (3D-cosd, $x^{\text{nor}}_{i}$, $x^{\text{nor}}_{j}$) & 7 & 92.8 \\
  E & (2D-Ed, $x'_{i} - x'_{j}$, $x'_{i}$, $x'_{j}$) & 10 & $\approx$ 92.2  \\
  \Xhline{0.8pt}
  \end{tabular}
  \label{Tab6:relation}
\end{table}

\vspace{12pt}
\noindent \textbf{Low-level relation ${\bm{\mathrm h}}$.}\,\,~The key to RS-CNN is learning from relation, thus how to define ${\bm{\mathrm h}}$ is an issue worth exploring. Actually, ${\bm{\mathrm h}}$ can be defined flexibly, as long as it could discriminatively reflect the underlying shape. To validate this claim and facilitate the understanding, we experiment with five intuitive relation definitions as examples, whose results are summarized in Table~\ref{Tab6:relation}.

As can be seen, using only 3D Euclidean distance as ${\bm{\mathrm h}}$, the accuracy can also reach 92.5\% (model A). This demonstrates the effectiveness of our RS-CNN for high-level geometric relation learning. Moreover, the performance is gradually improved with additional relation, including coordinates difference (model B) and coordinates themselves (model C). We also utilize the normal vectors of two points and their cosine distance as ${\bm{\mathrm h}}$, the result (model D) is 92.8\%. This indicates RS-CNN is also able to abstract shape information from the relation in normals.

Intuitively, the relation among points in the 2D view of point cloud can also reflect the underlying shape. Therefore, to validate our RS-CNN for shape abstraction on 2D relation, we forcibly set the value of one dimension in 3D coordinates to be zero, \textit{i.e.}, projecting 3D points onto a 2D plane of $\mathrm{XY}$, $\mathrm{XZ}$ and $\mathrm{YZ}$. The results are all around 92.2\% (model E), which is quite impressive. This further verifies the effectiveness of the proposed relation learning method.

\vspace{10pt}
\noindent \textbf{Robustness to point permutation and rigid transformation.} We compare the robustness of our RS-CNN with PointNet~\cite{c1_pointnet} and PointNet++~\cite{c2_pointnet2}. Note that all the models are trained without related data augmentations, \textit{e.g.}, translation or rotation, to avoid confusion in this test. In addition, although relation learning in RS-CNN is robust to rotation, the initial input features of 3D coordinates are affected. We address this issue by normalizing each sampled point subset to corresponding local coordinate system, which is determined by each sampled point and its normal. For a fair comparison, we also perform this normalization for PointNet++, as it learns over local subsets as well. The 3D Euclidean distance is applied as geometric relation ${\bm{\mathrm h}}$ in RS-CNN for this test. Table~\ref{Tab7:robust_invariant} summarizes the test results.

As can be seen, all the methods are invariant to permutation. However, PointNet is vulnerable to both translation and rotation while PointNet++ is sensitive to rotation. By contrast, our RS-CNN with geometric relation learning is invariant to these perturbations, making it powerful for robust shape recognition.

\begin{table}[t]
  \centering
  \footnotesize
  \caption{Robustness to point permutation and rigid transformation (\%). During testing, we perform random permutation (perm.) of points, add a small translation of $\pm$0.2 and counterclockwise rotate the input point cloud by 90$^{\circ}$ and 180$^{\circ}$ around $\mathrm{Y}$ axis.}
  \begin{threeparttable}
  \begin{tabular}{l|c|ccccc}
  \Xhline{0.8pt}
  method & acc. & perm. & +0.2 & -0.2 & 90$^{\circ}$ & 180$^{\circ}$ \\
  \Xhline{0.5pt}
  PointNet~\cite{c1_pointnet} & 88.7 & 88.7 & 70.8 & 70.6 & 42.5 & 38.6 \\
  PointNet++~\cite{c2_pointnet2} & 88.2$^{\dag}$ & 88.2 & 88.2 & 88.2 & 47.9 & 39.7 \\
  \textbf{Ours} & \textbf{90.3}$^{\dag}$ & \textbf{90.3} & \textbf{90.3} & \textbf{90.3} & \textbf{90.3} & \textbf{90.3} \\
  \Xhline{0.8pt}
  \end{tabular}
  \begin{tablenotes}
        \footnotesize
        \item[$\dag$]The accuracy drops a lot mainly because the forcible normalization of each local point subset could bring difficulty for shape recognition.
  \end{tablenotes}
  \end{threeparttable}
  \label{Tab7:robust_invariant}
\end{table}

\vspace{7pt}
\subsection{Visualization and Complexity Analysis}
\begin{figure}[t]
\centerline{\includegraphics[width=8.4cm]{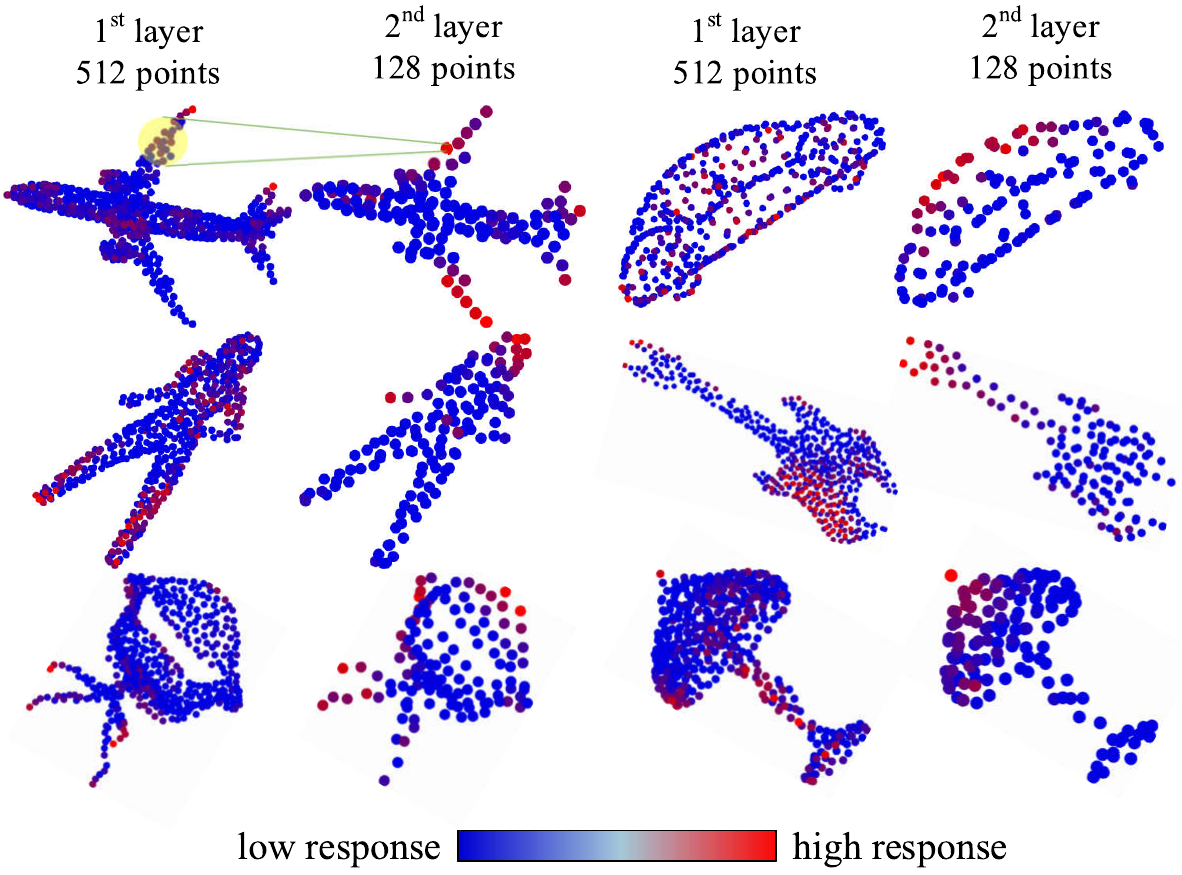}}
\caption{Visualization of the shape features learned by the first two layers of RS-CNN on ModelNet40 dataset. The features learned by the first layer mostly respond to edges, corners and arcs, while the ones in the second layer capture more semantical shape parts like airfoils and heads.}
\label{fig8:feature_show}
\vspace{7pt}
\end{figure}

\vspace{7pt}
\label{subsec4.3}
\noindent \textbf{Visualization.}\,\,~Fig.~\ref{fig8:feature_show} visualizes the shape features learned by the first two layers of RS-CNN on ModelNet40 dataset. As it shows, the features learned by the first layer mostly respond to edges, corners and arcs, while the ones in the second layer capture more semantical shape parts like airfoils and heads. This verifies RS-CNN can learn progressive shape-aware representation for point cloud analysis.

\vspace{10pt}
\noindent \textbf{Complexity Analysis.}\,\,~Table~\ref{Tab8:complexity} summarizes the space (number of params) and the time (floating point operations/sample) complexity of RS-CNN in classification with 1024 points as the input. Compared with PointNet~\cite{c1_pointnet}, RS-CNN reduces the params by 59.7\% and the FLOPs by 32.9\%, which shows its great potential for real-time applications, \textit{e.g.}, scene parsing in autonomous driving.

\begin{table}[t]
  \centering
  \caption{Complexity of RS-CNN in point cloud classification.}
  \begin{tabular}{l|cc}
  \Xhline{0.8pt}
  method & \#params & \#FLOPs/sample \\
  \Xhline{0.5pt}
  PointNet~\cite{c1_pointnet} & 3.50M & 440M \\
  PointNet++~\cite{c27} & 1.48M & 1684M \\
  PCNN~\cite{c27} & 8.20M & \textbf{294M} \\
  \textbf{Ours} & \textbf{1.41M} & 295M \\
  \Xhline{0.8pt}
  \end{tabular}
  \label{Tab8:complexity}
\end{table}

\vspace{10pt}
\section{Conclusion}
\label{sec:conclusion}
\vspace{10pt}
In this work, RS-CNN, namely, Relation-Shape Convolutional Neural Network, which extends regular grid CNN to irregular configuration for point cloud analysis, has been proposed.
The core to RS-CNN is a novel convolution operator, which learns from relation, \textit{i.e.}, the geometric topology constraint among points.
In this way, explicit reasoning about the spatial layout of points can be made to obtain discriminative shape awareness.
Moreover, the decent properties of geometric relation can also be acquired, such as robustness to rigid transformation.
As a consequence, RS-CNN equipped with this operator can achieve contextual shape-aware learning, making it highly effective.
Extensive experiments on challenging benchmarks across three tasks, as well as thorough empirical and theoretical analysis, have demonstrated RS-CNN achieves the state of the arts.

%

\newpage
{\small
\bibliographystyle{ieee}
\bibliography{egbib}

\begin{thebibliography}{10}\itemsep=-1pt

\bibitem{c16_eocnn}
M.~Atzmon, H.~Maron, and Y.~Lipman.
\newblock Point convolutional neural networks by extension operators.
\newblock In {\em SIGGRAPH}, pages 1--14, 2018.

\bibitem{c37}
Y.~Feng, Z.~Zhang, X.~Zhao, R.~Ji, and Y.~Gao.
\newblock {GVCNN}: Group-view convolutional neural networks for {3D} shape
  recognition.
\newblock In {\em CVPR}, pages 264--272, 2018.

\bibitem{c45}
M.~Gadelha, R.~Wang, and S.~Maji.
\newblock Multiresolution tree networks for {3D} point cloud processing.
\newblock In {\em ECCV}, pages 105--122, 2018.

\bibitem{flex_conv}
F.~Groh, P.~Wieschollek, and H.~P. Lensch.
\newblock Flex-convolution (million-scale point-cloud learning beyond
  grid-worlds).
\newblock {\em arXiv preprint arXiv:1803.07289}, 2018.

\bibitem{PCPNet}
P.~Guerrero, Y.~Kleiman, M.~Ovsjanikov, and N.~J. Mitra.
\newblock {PCPNet}: Learning local shape properties from raw point clouds.
\newblock {\em Comput. Graph. Forum}, 37(2):75--85, 2018.

\bibitem{multiview2}
H.~Guo, J.~Wang, Y.~Gao, J.~Li, and H.~Lu.
\newblock Multi-view {3D} object retrieval with deep embedding network.
\newblock {\em {IEEE} Trans. Image Processing}, 25(12):5526--5537, 2016.

\bibitem{c49}
Z.~Han, M.~Shang, Z.~Liu, C.~Vong, Y.~Liu, M.~Zwicker, J.~Han, and C.~L.~P.
  Chen.
\newblock {SeqViews2SeqLabels}: Learning {3D} global features via aggregating
  sequential views by {RNN} with attention.
\newblock {\em {IEEE} Trans. Image Processing}, 28(2):658--672, 2019.

\bibitem{conf_iccv_HeZRS15}
K.~He, X.~Zhang, S.~Ren, and J.~Sun.
\newblock Delving deep into rectifiers: {Surpassing} human-level performance on
  {ImageNet} classification.
\newblock In {\em ICCV}, pages 1026--1034, 2015.

\bibitem{c50}
K.~Hornik.
\newblock Approximation capabilities of multilayer feedforward networks.
\newblock {\em Neural Networks}, 4(2):251--257, 1991.

\bibitem{relation_detection}
H.~Hu, J.~Gu, Z.~Zhang, J.~Dai, and Y.~Wei.
\newblock Relation networks for object detection.
\newblock In {\em CVPR}, pages 3588--3597, 2018.

\bibitem{c17_pointwise}
B.-S. Hua, M.-K. Tran, and S.-K. Yeung.
\newblock Pointwise convolutional neural networks.
\newblock In {\em CVPR}, pages 974--993, 2018.

\bibitem{C28}
Q.~Huang, W.~Wang, and U.~Neumann.
\newblock Recurrent slice networks for {3D} segmentation of point clouds.
\newblock In {\em CVPR}, pages 2626--2635, 2018.

\bibitem{BN}
S.~Ioffe and C.~Szegedy.
\newblock Batch normalization: Accelerating deep network training by reducing
  internal covariate shift.
\newblock In {\em ICML}, pages 448--456, 2015.

\bibitem{stn}
M.~Jaderberg, K.~Simonyan, A.~Zisserman, and K.~Kavukcuoglu.
\newblock Spatial transformer networks.
\newblock In {\em NeurIPS}, pages 2017--2025, 2015.

\bibitem{bcl}
V.~Jampani, M.~Kiefel, and P.~V. Gehler.
\newblock Learning sparse high dimensional filters: Image filtering, dense
  {CRF}s and bilateral neural networks.
\newblock In {\em CVPR}, pages 4452--4461, 2016.

\bibitem{dfn_net}
X.~Jia, B.~D. Brabandere, T.~Tuytelaars, and L.~V. Gool.
\newblock Dynamic filter networks.
\newblock In {\em NeurIPS}, pages 667--675, 2016.

\bibitem{c23}
M.~Jiang, Y.~Wu, and C.~Lu.
\newblock {PointSIFT}: {A} {SIFT}-like network module for {3D} point cloud
  semantic segmentation.
\newblock {\em arXiv preprint arXiv:1807.00652}, 2018.

\bibitem{c26}
R.~Klokov and V.~S. Lempitsky.
\newblock Escape from cells: Deep {Kd-Networks} for the recognition of {3D}
  point cloud models.
\newblock In {\em ICCV}, pages 863--872, 2017.

\bibitem{alexnet}
A.~Krizhevsky, I.~Sutskever, and G.~E. Hinton.
\newblock {ImageNet} classification with deep convolutional neural networks.
\newblock In {\em NeurIPS}, pages 1106--1114, 2012.

\bibitem{c8_superpoint}
L.~Landrieu and M.~Simonovsky.
\newblock Large-scale point cloud semantic segmentation with superpoint graphs.
\newblock In {\em CVPR}, pages 4558--4567, 2018.

\bibitem{c20}
J.~Li, B.~M. Chen, and G.~H. Lee.
\newblock {SO-Net}: Self-organizing network for point cloud analysis.
\newblock In {\em CVPR}, pages 9397--9406, 2018.

\bibitem{c30}
R.~Li, S.~Wang, F.~Zhu, and J.~Huang.
\newblock Adaptive graph convolutional neural networks.
\newblock In {\em AAAI}, pages 3546--3553, 2018.

\bibitem{c27}
Y.~Li, R.~Bu, M.~Sun, and B.~Chen.
\newblock Point{CNN}: Convolution on {X}-transformed points.
\newblock In {\em NeurIPS}, pages 828--838, 2018.

\bibitem{vox2}
D.~Maturana and S.~Scherer.
\newblock {VoxNet}: {A} {3D} convolutional neural network for real-time object
  recognition.
\newblock In {\em IROS}, pages 922--928, 2015.

\bibitem{relu}
V.~Nair and G.~E. Hinton.
\newblock Rectified linear units improve restricted boltzmann machines.
\newblock In {\em ICML}, pages 807--814, 2010.

\bibitem{c1_pointnet}
C.~R. Qi, H.~Su, K.~Mo, and L.~J. Guibas.
\newblock {PointNet}: Deep learning on point sets for {3D} classification and
  segmentation.
\newblock In {\em CVPR}, pages 77--85, 2016.

\bibitem{c51}
C.~R. Qi, H.~Su, M.~Nie{\ss}ner, A.~Dai, M.~Yan, and L.~J. Guibas.
\newblock Volumetric and multi-view {CNN}s for object classification on {3D}
  data.
\newblock In {\em CVPR}, pages 5648--5656, 2016.

\bibitem{c2_pointnet2}
C.~R. Qi, L.~Yi, H.~su, and L.~J. Guibas.
\newblock {PointNet++}: Deep hierarchical feature learning on point sets in a
  metric space.
\newblock In {\em NeurIPS}, pages 5099--5108, 2017.

\bibitem{c6_synccnn}
S.~Ravanbakhsh, J.~Schneider, and B.~Poczos.
\newblock Deep learning with sets and point clouds.
\newblock In {\em ICLR}, pages 1--12, 2017.

\bibitem{c53}
G.~Riegler, A.~O. Ulusoy, and A.~Geiger.
\newblock {OctNet}: Learning deep {3D} representations at high resolutions.
\newblock In {\em CVPR}, pages 6620--6629, 2017.

\bibitem{c5_generalcnn}
A.~Savchenkov.
\newblock Generalized convolutional neural networks for point cloud data.
\newblock In {\em ICMLA}, pages 930--935, 2017.

\bibitem{c9_kcnet}
Y.~Shen, C.~Feng, Y.~Yang, and D.~Tian.
\newblock Mining point cloud local structures by kernel correlation and graph
  pooling.
\newblock In {\em CVPR}, pages 4548--4557, 2018.

\bibitem{c32}
M.~Simonovsky and N.~Komodakis.
\newblock Dynamic edge-conditioned filters in convolutional neural networks on
  graphs.
\newblock In {\em CVPR}, pages 29--38, 2017.

\bibitem{VGG}
K.~Simonyan and A.~Zisserman.
\newblock Very deep convolutional networks for large-scale image recognition.
\newblock In {\em ICLR}, pages 1--14, 2015.

\bibitem{dropout}
N.~Srivastava, G.~E. Hinton, A.~Krizhevsky, I.~Sutskever, and R.~Salakhutdinov.
\newblock {Dropout}: {A} simple way to prevent neural networks from
  overfitting.
\newblock {\em Journal of Machine Learning Research.}, 15(1):1929--1958, 2014.

\bibitem{c10_splatnet}
H.~Su, V.~Jampani, D.~Sun, S.~Maji, E.~Kalogerakis, M.-H. Yang, and J.~Kautz.
\newblock {SPLATNet}: Sparse lattice networks for point cloud processing.
\newblock In {\em CVPR}, pages 2530--2539, 2018.

\bibitem{multiview1}
H.~Su, S.~Maji, E.~Kalogerakis, and E.~G. Learned{-}Miller.
\newblock Multi-view convolutional neural networks for {3D} shape recognition.
\newblock In {\em ICCV}, pages 945--953, 2015.

\bibitem{vox3}
M.~Tatarchenko, A.~Dosovitskiy, and T.~Brox.
\newblock Octree generating networks: Efficient convolutional architectures for
  high-resolution {3D} outputs.
\newblock In {\em ICCV}, pages 2107--2115, 2017.

\bibitem{c19}
G.~Te, W.~Hu, A.~Zheng, and Z.~Guo.
\newblock {RGCNN}: Regularized graph {CNN} for point cloud segmentation.
\newblock In {\em MM}, pages 746--754, 2018.

\bibitem{c14_scn}
C.~Wang, B.~Samari, and K.~Siddiqi.
\newblock Local spectral graph convolution for point set feature learning.
\newblock In {\em ECCV}, pages 1--16, 2018.

\bibitem{c29}
P.~Wang, Y.~Liu, Y.~Guo, C.~Sun, and X.~Tong.
\newblock {O-CNN:} octree-based convolutional neural networks for {3D} shape
  analysis.
\newblock {\em {ACM} Trans. Graph.}, 36(4):72:1--72:11, 2017.

\bibitem{Param_conv}
S.~Wang, S.~Suo, W.~Ma, A.~Pokrovsky, and R.~Urtasun.
\newblock Deep parametric continuous convolutional neural networks.
\newblock In {\em CVPR}, pages 2589--2597, 2018.

\bibitem{non-local}
X.~Wang, R.~B. Girshick, A.~Gupta, and K.~He.
\newblock Non-local neural networks.
\newblock In {\em CVPR}, pages 7794--7803, 2018.

\bibitem{c22}
Y.~Wang, Y.~Sun, Z.~Liu, S.~E. Sarma, M.~M. Bronstein, and J.~M. Solomon.
\newblock Dynamic graph {CNN} for learning on point clouds.
\newblock {\em arXiv preprint arXiv:1801.07829}, 2018.

\bibitem{modelnet40}
Z.~Wu, S.~Song, A.~Khosla, F.~Yu, L.~Zhang, X.~Tang, and J.~Xiao.
\newblock {3D ShapeNets}: {A} deep representation for volumetric shapes.
\newblock In {\em CVPR}, pages 1912--1920, 2015.

\bibitem{multiview3}
J.~Xie, G.~Dai, F.~Zhu, E.~K. Wong, and Y.~Fang.
\newblock {DeepShape}: Deep-learned shape descriptor for {3D} shape retrieval.
\newblock {\em {IEEE} Trans. Pattern Anal. Mach. Intell.}, 39(7):1335--1345,
  2017.

\bibitem{c7_attsp}
S.~Xie, S.~Liu, Z.~Chen, and Z.~Tu.
\newblock Attentional {ShapeContextNet} for point cloud recognition.
\newblock In {\em CVPR}, pages 4606--4615, 2018.

\bibitem{c21}
Y.~Xu, T.~Fan, M.~Xu, L.~Zeng, and Y.~Qiao.
\newblock Spider{CNN}: Deep learning on point sets with parameterized
  convolutional filters.
\newblock In {\em ECCV}, pages 90--105, 2018.

\bibitem{c54}
L.~Yi, V.~G. Kim, D.~Ceylan, I.~Shen, M.~Yan, H.~Su, C.~Lu, Q.~Huang,
  A.~Sheffer, and L.~J. Guibas.
\newblock A scalable active framework for region annotation in {3D} shape
  collections.
\newblock {\em {ACM} Trans. Graph.}, 35(6):210:1--210:12, 2016.

\bibitem{c3_synccnn}
L.~Yi, H.~Su, X.~Guo, and L.~J. Guibas.
\newblock {SyncSpecCNN}: Synchronized spectral {CNN} for {3D} shape
  segmentation.
\newblock In {\em CVPR}, pages 6584--6592, 2017.

\bibitem{c24}
M.~Zaheer, S.~Kottur, S.~Ravanbakhsh, B.~P{\'{o}}czos, R.~R. Salakhutdinov, and
  A.~J. Smola.
\newblock Deep sets.
\newblock In {\em NeurIPS}, pages 3394--3404, 2017.

\bibitem{visualize}
M.~D. Zeiler and R.~Fergus.
\newblock Visualizing and understanding convolutional networks.
\newblock In {\em ECCV}, pages 818--833, 2014.

\end{thebibliography}
}

\newpage
\renewcommand\thesection{\Alph{section}}
\setcounter{section}{0}
{\Large \hspace{0.9cm} \textbf{Supplementary Material}}

\section{Outline}
\label{sec:Outline}
This supplementary material provides further investigations for the proposed RS-CNN. Specifically, three issues on the construction of local neighborhood are discussed in Sec \ref{sec:Neighborhood}. More details of the relation learning on 2D views of 3D point cloud are presented in Sec \ref{sec:Relation}. All the experiments are conducted on ModelNet40 dataset. 

\section{Construction of Local Neighborhood}
\label{sec:Neighborhood}
In the main paper, the local point subset $P_{\text{sub}}$ in Eq. \eqref{Eq3:graph_relation} is modeled to be a spherical neighborhood with a sampled point $x_{i}$ as the centroid, and the surrounding points as its neighbors $\mathcal{N}(x_{i})$ (see the left-most part in Fig. \ref{fig2:RS-GC}). Then, the inductive representation ${\bm{\mathrm f}}_{P_{\text{sub}}}$, which is expected to reason the spatial layout of points in this neighborhood, is obtained by performing the proposed relation-shape convolution to aggregate all the relation between $x_{i}$ and $\mathcal{N}(x_{i})$.

In the above process, there are mainly three issues worth further investigation: (1) How should $\mathcal{N}(x_{i})$ be selected? (2) Is it suitable to simply aggregate all the relation between $x_{i}$ and $\mathcal{N}(x_{i})$? (3) Is it reasonable to select the sampled point $x_{i}$ as the centroid? They are explored as follows.

\vspace{10pt}
\noindent \textbf{(1) Selection of the neighbors $\mathcal{N}(x_{i})$.} \ Two strategies, k-nearest neighbor (k-NN) and random picking in the ball (Random-PIB), are investigated for this issue. Table \ref{Tab1:knn} summarizes the results. Note that the number of neighbors is set to be equal for a fair comparison. As it shows, Random-PIB obtains better classification accuracy. The reason may be k-NN would suffer selection inhomogeneity in some cases, which is adverse to shape-aware learning (the aggregated relation may only focus on dense points and ignore sparse points that are essential for the underlying shape). By contrast, Random-PIB can have a better coverage of points even in the case of inhomogeneous distribution.

\vspace{10pt}
\noindent \textbf{(2) Relation aggregation issue.} \ To verify this issue, we randomly cut off some relation between $x_{i}$ and $\mathcal{N}(x_{i})$ during training, \textit{i.e.}, randomly setting the learned high-level relation expression $\mathcal{M}({\bm{\mathrm h}}_{ij})$ in Eq. \eqref{Eq3:graph_relation} to be a zero vector, but using all the relation during testing. This operation is similar to the dropout technique. Table \ref{Tab2:cut_relation} summarizes the results. As can be seen, the best approach is training with all the relation while the second best one is training with relation cut ratio of 0.3. This indicates the dropout-like technique is not suitable for relation learning, probably because RS-CNN can automatically encode the strength of the relation in the learning process.

\begin{table}[t]
  \centering
  \caption{The results (\%) of two selection strategies on $\mathcal{N}(x_{i})$. Both of them are trained with a single-scale neighborhood. For a fair comparison, the number of neighbors is set to be equal in each layer between the two models.}
  \begin{tabular}{l|c}
  \Xhline{0.8pt}
  method & acc. \\
  \Xhline{0.5pt}
  k-NN  & 90.5 \\
  Random-PIB & 92.2 \\
  \Xhline{0.8pt}
  \end{tabular}
  \label{Tab1:knn}
  \vspace{6pt}
\end{table}

\begin{table}[t]
  \centering
  \caption{The results (\%) of learning with relation in different proportions. ``ratio'' indicates the cut off relation accounts for the proportion of all the relation between the centroid and the neighbors.}
  \begin{tabular}{l|cccccc}
  \Xhline{0.8pt}
  ratio & 0 & 0.1 & 0.2 & 0.3 & 0.4 & 0.5 \\
  \hline
  acc. & 93.6 & 92.8 & 92.9 & 93.2 & 92.5 & 92.1 \\
  \Xhline{0.8pt}
  \end{tabular}
  \label{Tab2:cut_relation}
  \vspace{6pt}
\end{table}

\begin{table}[t]
  \centering
  \caption{The results (\%) of three selection approaches and one fusion strategy of the centroid. The approach of picking in $\mathcal{N}(x_{i})$ is performed randomly in each neighborhood. Note that the weight in $\mathcal{M}$ is shared over these approaches in the fusion process.}
  \begin{tabular}{l|cc}
  \Xhline{0.8pt}
  centroid & acc. \\
  \Xhline{0.5pt}
  sampled point $x_{i}$ & 93.6 \\
  average of $\mathcal{N}(x_{i})$ & 93.6 \\
  random picking in $\mathcal{N}(x_{i})$ & 92.8 \\
  fusion of above & 93.4 \\
  \Xhline{0.8pt}
  \end{tabular}
  \label{Tab3:centroid}
  \vspace{6pt}
\end{table}

\begin{table}[t]
  \centering
  \caption{The results (\%) of RS-CNN with the low-level relation ${\bm{\mathrm h}}$ defined on 2D views ($\mathrm{XY}$-Ed: Euclidean distance in $\mathrm{XY}$ plane, $x^{\mathrm{xy}}$: 2D coordinates of $x$ in $\mathrm{XY}$ plane, \textit{i.e.}, the value of $\mathrm{z}$ is set to be zero). The fusion strategy is achieved by performing element-wise summation of ${\bm{\mathrm f}}_{P_{\text{sub}}}$ in Eq. \eqref{Eq3:graph_relation}, with ${\bm{\mathrm h}}$ defined on three 2D views. Note that the weight in $\mathcal{M}$ is shared over these three views in the fusion process.}
  \begin{tabular}{lc|c}
  \Xhline{0.8pt}
  low-level relation ${\bm{\mathrm h}}$ & channels & acc. \\
  \Xhline{0.5pt}
  ($\mathrm{XY}$-Ed, $x^{\mathrm{xy}}_{i} - x^{\mathrm{xy}}_{j}$, $x^{\mathrm{xy}}_{i}$, $x^{\mathrm{xy}}_{j}$)  & 10 & 92.1 \\
  ($\mathrm{XZ}$-Ed, $x^{\mathrm{xz}}_{i} - x^{\mathrm{xz}}_{j}$, $x^{\mathrm{xz}}_{i}$, $x^{\mathrm{xz}}_{j}$)  & 10 & 92.1 \\
  ($\mathrm{YZ}$-Ed, $x^{\mathrm{yz}}_{i} - x^{\mathrm{yz}}_{j}$, $x^{\mathrm{yz}}_{i}$, $x^{\mathrm{yz}}_{j}$)  & 10 & 92.2 \\
  fusion of above three views &  & 92.5 \\
  \Xhline{0.8pt}
  \end{tabular}
  \label{Tab4:relation}
  \vspace{0pt}
\end{table}

\vspace{10pt}
\noindent \textbf{(3) Selection of the centroid.} \ Three types of the centroid: the sampled point $x_{i}$, the average of $\mathcal{N}(x_{i})$ and random picking in $\mathcal{N}(x_{i})$, are studied for this issue. Besides, a strategy that fuses all of them is also studied. The results are summarized in Table \ref{Tab3:centroid}, where the first two strategies obtain the same decent accuracy while random picking performs less well. The reason may be that random picking requires RS-CNN to reason the spatial layout of points from various topological connections, which is quite difficult.

Another promising strategy is fusing a group of relations that are centered on different centroids. This can be achieved by performing element-wise summation of ${\bm{\mathrm f}}_{P_{\text{sub}}}$ in Eq. \eqref{Eq3:graph_relation}, with the relation centered on the above three kinds of centroids. However, it does not perform better, with an accuracy of 93.4\% that is lower than the best single-centroid version of 93.6\%.

\section{Low-Level Relation ${\bm{\mathrm h}}$}
\label{sec:Relation}
\begin{figure}[t]
\centerline{\includegraphics[width=5.9cm]{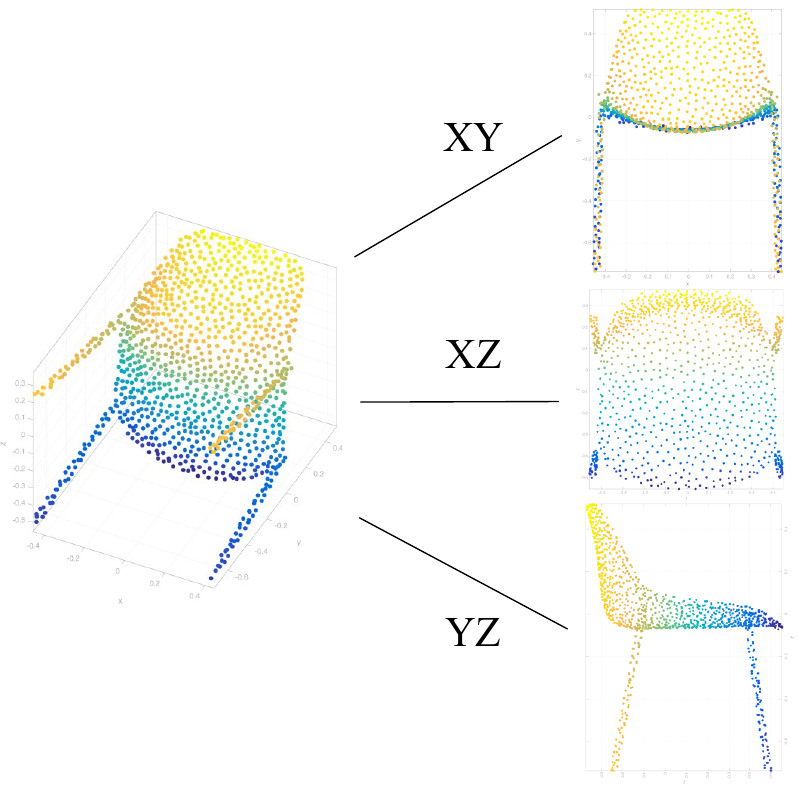}}
\caption{The projection of 3D point cloud onto the 2D plane of $\mathrm{XY}$, $\mathrm{XZ}$ and $\mathrm{YZ}$.}
\label{fig1:2d_relation}
\end{figure}

More details of the relation learning on 2D views of point cloud (the fourth part in Sec \ref{subsec4.2}) are provided in this section. As illustrated in Fig. \ref{fig1:2d_relation} in this material, the relation among points in the 2D view can also reflect the underlying shape. Therefore, we are interested in how powerfully the proposed RS-CNN to acquire shape awareness from only 2D-view relation of points.

To validate this, the value of one dimension in 3D coordinates is forcibly set to be zero, that is, 3D points are projected onto the 2D plane of $\mathrm{XY}$, $\mathrm{XZ}$ and $\mathrm{YZ}$ for three 2D views. In addition, a strategy with fusion of these views is also studied. Note that the projection operation is only conducted for the definition of ${\bm{\mathrm h}}$, the initial input features for $x_{j}$ in Eq. \eqref{Eq3:graph_relation} is still intact 3D coordinates. Table \ref{Tab4:relation} summarizes the results. As can be seen, all single-view relation can achieve an accuracy around 92.2\%, which is quite impressive. After fusing them, the result is improved by 0.3\%. This shows RS-CNN can also capture the underlying shape well even with relation learning from 2D view (potentially, a group of 2D views) of 3D point cloud, further verifying its effectiveness.

\end{document}